\documentclass[conference]{IEEEtran}
\IEEEoverridecommandlockouts
\usepackage{cite}
\usepackage{amsmath,amssymb,amsfonts}
\usepackage{algorithmic}
\usepackage{graphicx}
\usepackage{textcomp}
\usepackage{xcolor}
\usepackage{xspace}
\usepackage{hyperref} 
\hypersetup{
    colorlinks=true,       
    linkcolor=blue,        
    citecolor=blue,        
    filecolor=blue,        
    urlcolor=blue          
}
\usepackage{threeparttable}
\usepackage{booktabs}
\usepackage{multirow}
\usepackage{bm}
\usepackage{array}
\usepackage{makecell}
\usepackage{svg}
\usepackage[normalem]{ulem}
\def\BibTeX{{\rm B\kern-.05em{\sc i\kern-.025em b}\kern-.08em
    T\kern-.1667em\lower.7ex\hbox{E}\kern-.125emX}}

\IEEEoverridecommandlockouts

\def\0{{\mathbf 0}}
\def\1{{\mathbf 1}}

\def\E{{\mathbf E}}

\def\H{{\mathbf H}}

\def\T{{\mathbf T}}

\def\V{{\mathbf V}}

\def\Y{{\mathbf Y}}

\def\ie{{\textit{i.e.}}}
\def\eg{{\textit{e.g.}}}

\def\0{{\mathbf 0}}
\def\1{{\mathbf 1}}

\def\E{{\mathbf E}}

\def\H{{\mathbf H}}

\def\T{{\mathbf T}}

\def\V{{\mathbf V}}

\def\Y{{\mathbf Y}}

\def\ie{{\textit{i.e.}}}
\def\eg{{\textit{e.g.}}}

\newcommand{\ourmethod}{{SpatialGeo}\xspace}

\begin{document}

\title{SpatialGeo: Boosting Spatial Reasoning in Multimodal LLMs via Geometry-Semantics Fusion
}



\author{\IEEEauthorblockN{Jiajie Guo$^{\star}$, Qingpeng Zhu$^{\dagger}$, Jin Zeng$^{\star}$, Xiaolong Wu$^{\star}$, Changyong He$^{\star}$, Weida Wang$^{\star}$}
\IEEEauthorblockA{$^{\star}$School of Computer Science and Technology, Tongji University, Shanghai, China ~~~
$^{\dagger}$Independent Researcher
}
\thanks{The work was supported in part by the National Natural Science Foundation of China under Grant 62201389, and in part by the Fundamental Research Funds for the Central Universities under Grant 22120230311. \textit{Corresponding author: Jin Zeng (zengjin@tongji.edu.cn)}.}
}

\maketitle

\begin{abstract}
Multimodal large language models (MLLMs) have achieved significant progress in image and language tasks due to the strong reasoning capability of large language models (LLMs).
Nevertheless, most MLLMs suffer from limited spatial reasoning ability to interpret and infer spatial arrangements in three-dimensional space. 
In this work, we propose a novel vision encoder based on hierarchical fusion of geometry and semantics features, generating spatial-aware visual embedding and boosting the spatial grounding capability of MLLMs. 
Specifically, we first unveil that the spatial ambiguity shortcoming stems from the lossy embedding of the vision encoder utilized in most existing MLLMs (\eg, CLIP), restricted to instance-level semantic features.
This motivates us to complement CLIP with the geometry features from vision-only self-supervised learning via a hierarchical adapter, enhancing the spatial awareness in the proposed \ourmethod. 
The network is efficiently trained using pretrained LLaVA model and optimized with random feature dropping to avoid trivial solutions relying solely on the CLIP encoder.
Experimental results show that \ourmethod improves the accuracy in spatial reasoning tasks, enhancing state-of-the-art models by at least $8.0\%$ in SpatialRGPT-Bench with $\sim50\%$ less memory cost during inference.
The source code is available via \url{https://ricky-plus.github.io/SpatialGeoPages/}. 

\end{abstract}


\section{Introduction}
Due to their strong reasoning and generalization capabilities, large language models (LLMs) have led to remarkable progress across various language understanding and generation tasks, demonstrating human-level performance in tasks such as question answering, logical inference, and code generation \cite{llm2023survey}. 
However, to extend LLMs to real-world interaction scenarios, \eg, embodied agents, it is essential for the models to possess \textit{spatial reasoning ability} to comprehend, infer, and articulate the spatial attributes and relationships of objects within physical space \cite{hong20233d,zhu2024multi,shi2025multi,yin2025shapegpt}.
This requires not only language-based reasoning but also a grounded understanding of visual input from the three-dimensional (3D) scene \cite{Yang_2025_CVPR}.

\begin{figure}[t]
    \centering
    \includegraphics[width=0.85\linewidth]{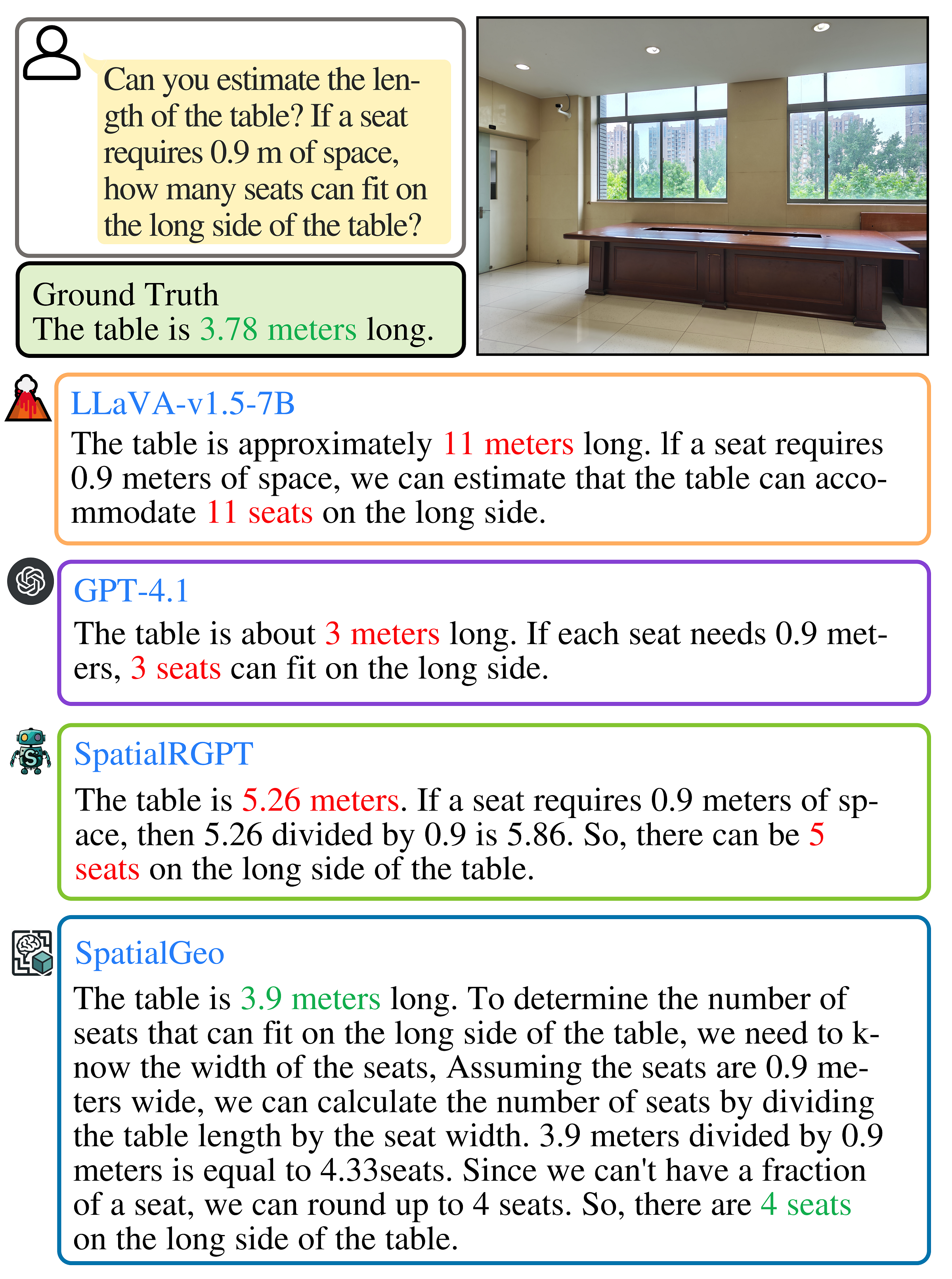}
    \vspace{-0.3cm}
    \caption{\ourmethod demonstrates superior performance in complex spatial reasoning tasks compared to SOTA models, including open-source LLaVA-v1.5 \cite{llava2024}, closed-source GPT-4.1 \cite{gpt4.1} and 3D-LLM SpatialRGPT \cite{cheng2025spatialrgpt}.}
    \label{fig:intro}
    \vspace{-0.5cm}
\end{figure}

Leveraging the reasoning ability of LLMs, recent efforts have led to the development of multimodal large language models (MLLMs) that combine LLMs with vision encoders to enable image-grounded reasoning \cite{survey_on_MLLMs,llava2023,gpt4.1}.
For instance, the state-of-the-art (SOTA) LLaVA \cite{llava2023}
uses pre-trained contrastive language image pre-training (CLIP) \cite{clip} as the vision encoder, which is integrated with pretrained LLMs (\eg, LLaMA \cite{touvron2023llama}) to enable tasks such as image captioning, visual question answering (VQA), \textit{etc}. 
However, existing MLLMs still suffer from visual shortcomings in performing spatial reasoning.
For example, as illustrated in Fig.\,\ref{fig:intro}, the SOTA LLaVA-v1.5-7B model \cite{llava2024} fails to infer the size of the table and cannot correctly answer for the seat placement.
Even the latest GPT-4.1 \cite{gpt4.1} does not provide precise spatial reasoning.
%
In sum, \textit{it remains a major challenge to enhance spatial reasoning capability of MLLMs}, which greatly limits their applications in critical tasks such as robotic navigation and manipulation, visual planning, \textit{etc} \cite{hong20233d,3dllmsruvey2024}.  

To tackle the above challenge, we propose \textbf{\ourmethod} which enhances spatial reasoning in MLLMs based on the novel vision encoder generating spatial-aware visual embedding.
First, we analyze the reason for the \textit{spatial ambiguity issue} in existing MLLMs.
%
Specifically, inspired by MMVP \cite{mmvp}, we examine 
the embedding space of the CLIP vision encoder.
It turns out that for a pair of images captured at different focal lengths of the same scene, CLIP encodes them into highly similar features, indicating a lossy spatial information encoding. 
Then the error in the vision encoder leads to downstream failures in LLMs, which explains the result of LLaVA in Fig.\,\ref{fig:intro} where the objects are correctly recognized but their spatial relations are erroneously inferred.
Therefore, we emphasize that \textit{the key to enhancing spatial reasoning capability lies in the spatial grounding capability of vision encoder in MLLM.}




Although there are existing attempts to improve the spatial grounding by injecting 3D features into MLLM encoder \cite{3dllmsruvey2024}, the performance is limited.
Some methods utilize prompt tuning for guiding LLMs toward 3D task comprehension in a training-free manner \cite{spatialpin,zhang2024agent3d}. 
For instance, SpatialPIN \cite{spatialpin} utilizes prior knowledge from 3D foundation models to facilitate spatial understanding via prompt engineering. 
However, \textit{the 3D features in the prompts are not guaranteed to be aligned with the text tokens, so the performance is limited.}
Another category of methods explicitly inputs the depth maps inferred from monocular images and includes extra feature extraction modules for depth images in MLLMs \cite{cheng2025spatialrgpt,cai2024spatialbot}. 
For example, SpatialRGPT \cite{cheng2025spatialrgpt} estimates depth maps from input images as auxiliary inputs and additionally introduces a region representation module to specify regions of objects in the question. 
Nevertheless, \textit{these schemes require the extra modules which may accumulate errors and limit the accuracy} as shown in Fig.\,\ref{fig:intro}.
Moreover, this leads to \textit{heavy computational overheads and poses difficulties in training and inference efficiency}.

In contrast, the proposed \ourmethod efficiently enhances spatial grounding as shown in Fig.\,\ref{fig:intro} by complementing \textit{instance-level semantic features} with \textit{geometry-aware visual embedding} in the vision encoder.
Specifically, since CLIP features suffer from spatial ambiguity, we involve the spatial-aware MoGe encoder \cite{moge}, a vision-only self-supervised model which is capable of recovering 3D geometry from monocular images and resolves the spatial ambiguity in the embedding space.
To effectively integrate MoGe, we design a hierarchical adapter that progressively fuses multiple layers of MoGe encoder generating a mixture of geometry and semantic features, which is then interleaved with CLIP features before being fed into LLM.
In addition, we develop an efficient training strategy using the pretrained LLaVA model \cite{llava2024} to promote training and data efficiency, then optimize the network with the add-on MoGe encoder using random feature dropping to avoid trivial solutions relying solely on CLIP.
Our main contributions are summarized as follows:



\begin{itemize}
    \item We unveil the spatial ambiguity issue in existing MLLMs and propose \ourmethod fusing geometry and semantics features in the vision encoder, boosting the spatial grounding capability of MLLMs;
    \item We design the hierarchical adapter to effectively fuse the mixture of geometry and semantic features from the vision-centric encoder;  
    \item We develop an efficient training strategy using pretrained LLaVA with the add-on MoGe and avoid trivial solutions via random feature dropping.
\end{itemize}
Extensive experiments show that \ourmethod improves the accuracy in spatial-related VQA tasks, enhancing SOTA models by at least $8.0\%$ in SpatialRGPT-Bench \cite{cheng2025spatialrgpt} with $\sim50\%$ less memory cost during inference.

\section{Preliminary}
\begin{figure*}[th]
    \centering
    \includegraphics[width=1.0\textwidth]{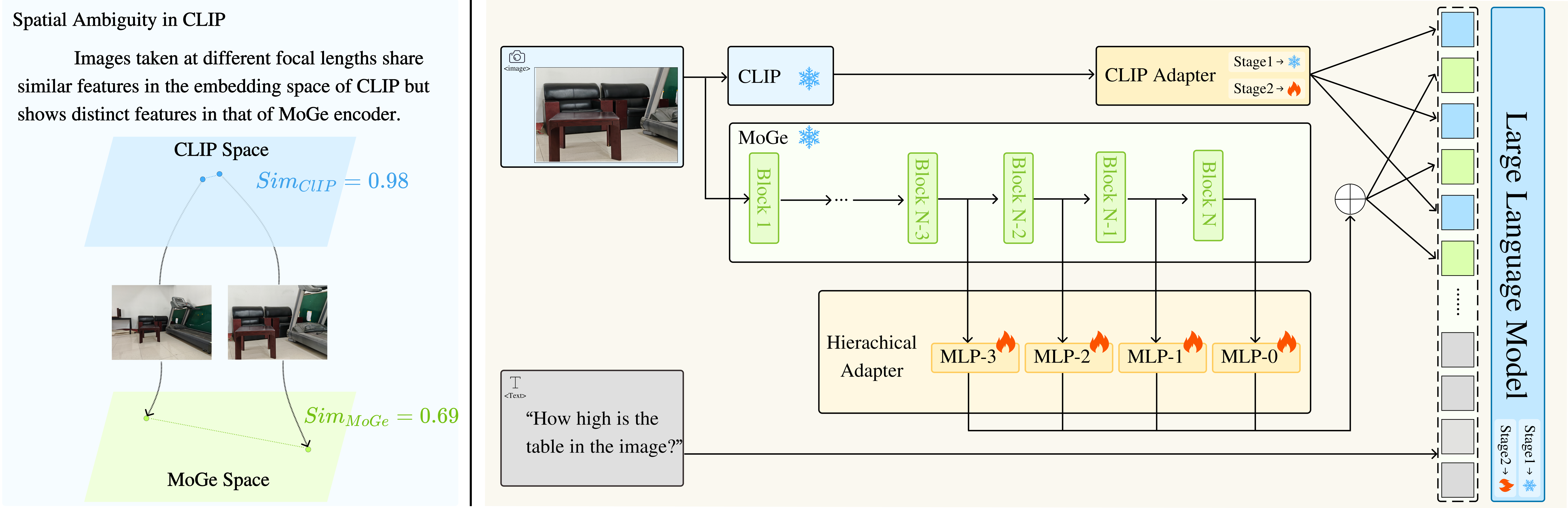}
    \caption{Left: The gap between CLIP and MoGe embedding space, where CLIP fails to distinguish images with different spatial information. 
    Right: Overall architecture of \ourmethod composed of CLIP module, MoGe module with hierarchical adapter, and LLM module with mixture of geometry and semantic features. The hierarchical adapter is first optimized with freezed vision encoders; then in instruction tuning, all adapters and LLM are updated with a random feature dropping strategy.}
    \label{fig:model_structure}
    \vspace{-0.5cm}
\end{figure*}

As a representative MLLM, LLaVA \cite{llava2023,llava2024} integrates a pre-trained vision encoder with an LLM to enable general-purpose visual and language understanding.
LLaVA consists of four modules: the visual encoder $f_\mathrm{V}$, the adapter $f_\mathrm{P}$, the input text embedding layer $f_\mathrm{T}$ and the LLM $f_\mathrm{L}$.
When the visual input $\V \in \mathbb{R}^{H \times W \times 3}$  and text input $\T$ are received, the visual encoder $f_\mathrm{V}$ of the model processes the visual input $\V$ to get visual features $\V'$ as
\begin{equation} \label{eq_encoder}
    \V'=f_{\mathrm{V}}(\V) \in \mathbb{R}^{N_\mathrm{V} \times D_\mathrm{V}},
\end{equation}
where $N_\mathrm{V}$ is the number of visual tokens, $D_\mathrm{V}$ is the visual feature dimension.
$\V'$ is then processed by the adapter $f_\mathrm{P}$ composed of two layers of multi-layer perceptron (MLP) and Gaussian Error Linear Unit (GELU) activation to obtain visual embeddings $\mathbf{E}_{\mathrm{V}}$ as
\begin{equation}\label{eq_adapter_v}
    \mathbf{E}_{\mathrm{V}} = f_{\mathrm{P}}(\V') =  \text{MLP}(\text{GELU}(\text{MLP}(\V')) \in \mathbb{R}^{N_\mathrm{V} \times D}, \\
\end{equation}
where $D$ is the embedding dimension.
Simultaneously, the model calculates text embeddings $\mathbf{E}_{\mathrm{T}}$ as, 
\begin{equation} \label{eq_adapter}
    \mathbf{E}_{\mathrm{T}}=f_{\mathrm{T}}(\T) \in \mathbb{R}^{N_\mathrm{T} \times D},
\end{equation}
where $N_\mathrm{T}$ is the number of text input tokens.
The generated embedding vectors $E_\mathrm{T}$ and $E_\mathrm{V}$ are spliced into a single token sequence $\H_{\text{in}}$ and feed it into LLM to obtain generate response $\Y = \{ y_1, y_2, \dots, y_M \}$ of length $M$ with conditional probability:
\begin{equation}
P(\Y \mid \V, \T) = \prod_{m=1}^{M} P(y_m \mid y_{[1:m-1]}, \H_{\text{in}}).
\end{equation}
The entire model except the visual encoder $f_\text{V}$ is trained via instruction tuning to maximize $P$.



Although LLaVA \cite{llava2024} achieves the SOTA performance in general-purpose VQA, it still has limitations in processing spatial reasoning tasks.
Moreover, existing experiment shows that, although the model's spatial reasoning capacity improves with the increase in parameter scale, its performance in absolute spatial metrics remains limited, even for the largest-scale models \cite{spatialscore}. 
To overcome these shortcomings, we propose \ourmethod, which significantly enhances the capability of MLLMs in complex spatial reasoning tasks via the novel geometry-semantic fusion in the vision encoder and the efficient training strategy.





\section{Proposed Network}
\label{sec:3_method}

In this section, we first analyze the reason for visual shortcomings in existing MLLMs in Sec.\,\ref{sec:3_blind-pair}, which motivates us to complement semantic features with geometry-aware embeddings in the vision encoder in the proposed \ourmethod in Sec.\,\ref{sec:3_all}.
Detailed design of hierarchical adapter for visual feature fusion and training strategy to enhance training and data efficiency is discussed in Sec.\,\ref{sec:3_adapter} and Sec.\,\ref{met_traing_strategy}, respectively.

\subsection{Analysis of Spatial Ambiguity in MLLMs}
\label{sec:3_blind-pair}


To investigate the reason for the visual shortcomings in existing MLLMs, we follow MMVP \cite{mmvp} and examine the embedding spaces of CLIP \cite{clip} used in the SOTA LLaVA \cite{llava2024}.
Specifically, we compare the gap between the embedding space of CLIP and MoGe encoder, where MoGe is a vision-only self-supervised learning model and specifically designed for precise 3D geometry reconstruction from monocular open-domain images \cite{moge}.
As shown in the left side of Fig.\,\ref{fig:model_structure}, we capture two images at different focal lengths of the same scene and encode the image pair with CLIP and MoGe, respectively, and then compute the cosine similarity between the features of the image pair.

It turns out CLIP features show high similarity of $0.98$, indicating that the two images are indistinguishable in CLIP embedding due to the lossy spatial information encoding.
This \textit{spatial ambiguity} shortcoming in CLIP leads to downstream errors in LLMs, which explains why MLLMs are limited in spatial VQA tasks as exemplified in Fig.\,\ref{fig:intro}.
In contrast, the image pair features show a distinct difference in the MoGe embedding generated from its last layer, showing a reduced similarity $0.69$ and indicating \textit{geometry awareness} of MoGe embedding.





\subsection{Overall Architecture of \ourmethod}
\label{sec:3_all}

To resolve the spatial ambiguity in existing MLLMs based on the above analysis, we proposed \ourmethod to enhance spatial reasoning ability by integrating geometry-aware MoGe encoder into CLIP restricted to semantic features.
The overall architecture of \ourmethod is illustrated in Fig.\,\ref{fig:model_structure}, which is composed of three major modules: 1) \textit{CLIP module} with the CLIP encoder and its adapter to extract instance-level semantic features; 2) \textit{MoGe module} with the MoGe encoder and its adapter to embed a mixture of geometry and semantic features; 3) \textit{LLM module} with interleaved geometry and semantic embeddings together with text tokens as inputs to generate question answering.



The network architecture adopts LLaVA-1.5-7B \cite{llava2024} as the foundational model with moderate model scale, and incorporates MoGe encoder \cite{moge} pretrained with DINOv2 \cite{oquab2024dinov2,dino}.
The questions remains are: 1) how to effectively fuse the geometry-semantics features; 2) how to efficiently train the network with reduced training and data requirements.
These issues are addressed as follows.


\subsection{Hierarchical Adapter for Feature Fusion}
\label{sec:3_adapter}


Existing 3D-LLMs \cite{cai2024spatialbot,cheng2025spatialrgpt} that explicitly utilize 3D representations (\eg, depth maps) as input to guide MLLMs for 3D comprehension, requiring extra geometry feature extraction and alignment modules, thus leading to heavy computational cost and data requirement.
In contrast, we design a hierarchical adapter appended to the MoGe encoder to progressively fuse geometry-semantic features. Since both features are generated from the same encoder, the geometry features are inherently aligned with semantics, which eases the alignment to CLIP and the subsequent alignment to text tokens in LLaVA foundation model.

\noindent \textbf{Hierarchical Adapter}
Different blocks in ViT in MoGe generate different characteristics of features.
Specifically, features at deeper blocks encapsulate more abstract global semantics that are effective for tasks such as image classification and recognition.
On the other hand, features at shallower blocks close to the input preserve the underlying fine-grained details and exhibit stronger spatial grounding capabilities.


Therefore, as shown in the Fig.\,~\ref{fig:model_structure}, we propose the hierarchical adapter that consists of four sub-adapters with structure defined in (\ref{eq_adapter_v}), which are used to weight the features of the last four blocks in the MoGe encoder, respectively.
The outputs from sub-adapters are then summed up to obtain the final output. 
As shown in (\ref{eq_adapts}), $f_\mathrm{{V_i}}(\V)$ represents the features produced by the $i$-th block and $f_{\mathrm{P}_{(N-i)}}$ represents the adapter corresponding to the $i$-th block. The final feature $\E_\mathrm{M}$ is computed as:
\begin{equation} \label{eq_adapts}
    \E_\mathrm{M}=\sum_{i=N-3}^{N}f_{\mathrm{P}_{(N-i)}}(f_{\mathrm{V_{i}}}(\V)),
\end{equation}
where $N$ is the number of blocks in the MoGe encoder.





\noindent \textbf{Interleaved Feature Fusion}
Features from CLIP and MoGe with the same dimension $N_\text{V} \times D$ are interleaved while maintaining their original spatial order \cite{mmvp} as shown on the right of Fig.\,\ref{fig:model_structure}.
The resultant interleaved feature sequence is subsequently fed into the LLM for processing.

In sum, with the proposed hierarchical adapter, \textit{the resulting network does not require complicated geometry feature alignment modules nor extra requirements for training data,} improving the effectiveness and efficiency in \ourmethod.

\subsection{Training Strategy for Efficient Optimization} 
\label{met_traing_strategy}

For efficient training, we use the pretrained model of LLaVA-1.5-7B \cite{llava2024} for initialization of CLIP and LLM modules.
This avoids training vision encoders from scratch which requires time-consuming training and large-scale datasets.
We adopt a two-stage training, where we first pretrain the hierarchical adapter for MoGe feature alignment, then fine-tune the adapters and LLM for instruction tuning.
For detailed training configuration, we maintain the original training parameter settings of LLaVA-1.5 \cite{llava2024}.

\begin{figure*}[th]
    \centering
    \includegraphics[width=1.0\textwidth]{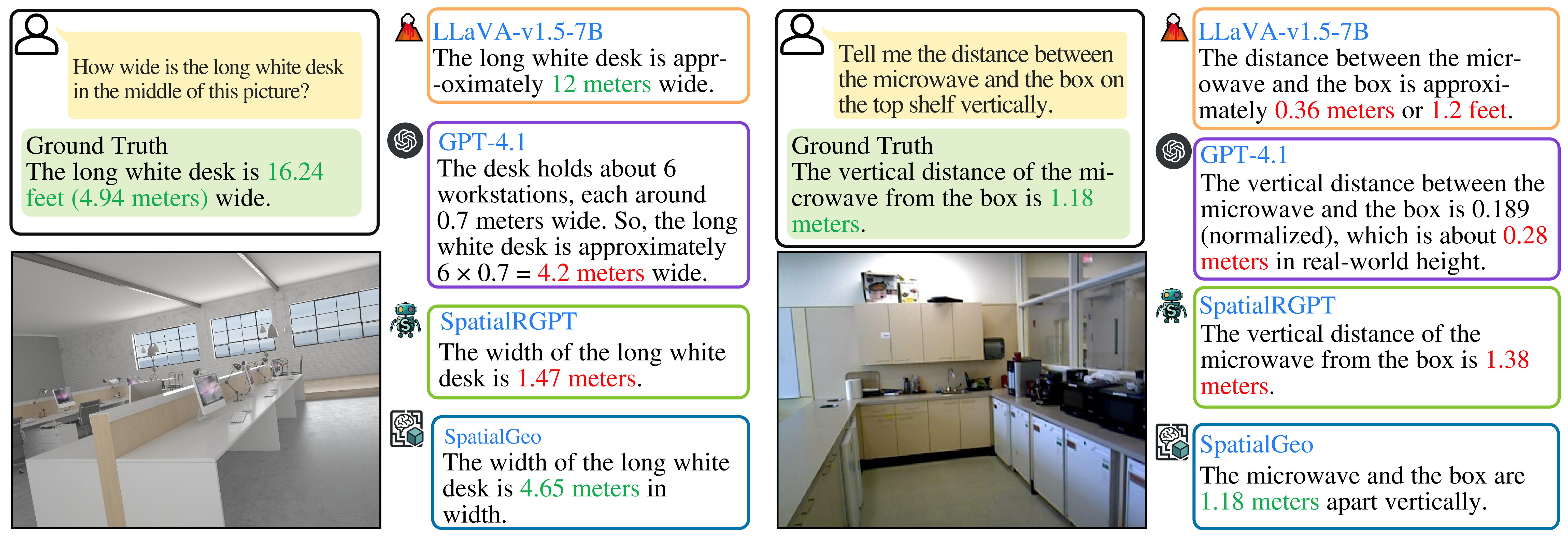}
    \vspace{-0.7cm}
    \caption{Sample VQA in SpatialRGPT-Bench \cite{cheng2025spatialrgpt}. Compared to LLaVA-v1.5-7B \cite{llava2024}, GPT-4.1 \cite{gpt4.1} and SpatialRGPT \cite{cheng2025spatialrgpt}, \ourmethod exhibits superior performance in quantitative spatial VQA tasks, \eg, inferring object sizes and relative distances.}
    \label{fig:exp}
    \vspace{-0.3cm}
\end{figure*}

\noindent \textbf{Stage 1: Pre-training for Features Alignment}
In the first stage, only the parameters of the hierarchical adapter are updated, while CLIP and its adapter are freezed. 
This preserves the feature alignment capability of CLIP adapter, 
and also enforces the network to fulfill the integration of MoGe.

\noindent \textbf{Stage 2: Instruction Fine-tuning with Random Feature Dropping}
In the second stage, we train the adapters of CLIP and MoGe, as well as the LLM with instruction data used in LLaVA-1.5 training and the spatial-related VQA dataset to be described in Sec.\,\ref{met_osd}. 
We adopt the LoRA fine-tuning strategy \cite{hu2022lora} with ZeRO-2 optimizer \cite{rajbhandari2020zero} for efficient training. 
Furthermore, we design a random feature dropping strategy that
masks the CLIP embeddings during training with a probability set to $0.3$.
This forces the LLM to focus on the utilization of MoGe features. 

\subsection{Spatial VQA Dataset Annotation Adjustment} 
\label{met_osd}


For instruction tuning, to enhance the network's ability to comprehend spatial relationships, we adopt the Open Spatial Dataset (OSD) \cite{cheng2025spatialrgpt} as part of the training set, which has rich and accurate spatial annotations.
We randomly extracted $160$k images from the total $\sim900$k images in OSD for training efficiency. 
In addition, the bounding box annotations in OSD are based on the original image dimension, while LLaVA-1.5 involves the resizing of input images to a fixed size of $\mathrm{336 \times 336}$ before putting them into the network.
Therefore, we adjust the bounding box annotations to be specified with the relative positions in the images so as to be used in more general MLLM training settings. 


\section{Experimental Results}
We first provide training details of \ourmethod then show comparison results with SOTA MLLMs datasets including spatial VQA benchmarks and the general VQA benchmarks.

\subsection{Experimental Setup}
\noindent \textbf{Dataset}
In the first stage of training, we employ the same dataset utilized in the first stage of LLaVA-1.5, namely the LLaVA-Pretrain dataset.
In the second stage of training, we select two datasets from the second-stage training of LLaVA-1.5 that are most relevant to our tasks, \ie, COCO and part one of VisualGenome \cite{llava2024}, and the adjusted OSD discussed in Sec.\,\ref{met_osd}.

\noindent \textbf{Training Setup}
Our training strategy is described in detail in Sec.\,\ref{met_traing_strategy}.
The training is conducted on Intel Xeon Platinum 8336C CPU@2.30GHz and 8 A800 GPUs for 8 hours for the first stage, and 17 hours for the second stage.

\subsection{Comparison with Existing Schemes for Spatial VQA} 
\label{exp_Comparative}
We compare our \ourmethod with SOTA MLLMs on spatial VQA datasets, including SpatialRGPT-Bench \cite{cheng2025spatialrgpt} which is the testing dataset in OSD, and SpatialScore \cite{spatialscore} to evaluate the generalization to unseen dataset.


\noindent \textbf{SpatialRGPT-Bench}
We adopt 642 quantitative VQA samples from SpatialRGPT-Bench for testing with five different types of questions shown in Table \ref{tab:Srgpt}.
The original strategy for evaluation is to submit the answers to GPT-4 for judging. 
However, given the diversity of the generated answers, this evaluation strategy can introduce error. 
Therefore, we refine evaluation method where we extract the numerical values and the numerical unit next to the numerical value in the answers and then convert the values to equivalent values in ``meters''.
The ground truth answers are processed in the same way. 
If the generated answer falls between $0.75$ and $1.25$ of the ground truth, the answer is considered correct.

We compare with SOTA MLLMs, including general-purpose LLaVA-1.5-7B \cite{llava2024}, GPT-4o, GPT-4V, the latest GPT-4.1 \cite{gpt4.1}, and the MLLM with 3D input, \ie, SpatialRGPT \cite{cheng2025spatialrgpt}.
For fair comparison, we finetune the open-source LLaVA-1.5-7B with the same dataset used in our second-stage training, namely LLaVA-OSD; for closed-source GPT models, the finetuning cannot be implemented. 
The results are shown in Table \ref{tab:Srgpt} where \ourmethod outperforms all the competing models in terms of the average accuracy, enhancing SpatialRGPT with the second-best results by at least \textbf{8.0\%} despite that SpatialRGPT is trained with the full OSD dataset, while we only use one-fifth of the original dataset.
Moreover, due to the simplified model structure, the memory cost during inference is reduced by $\sim$\textbf{50\%} compared with SpatialRGPT, validating the efficiency of \ourmethod.

In addition, we demonstrate sample VQA in Fig.\,\ref{fig:exp} to show spatial understanding tasks for inferring object sizes, relative distance between objects, where \ourmethod identifies the precise metric values.
More demonstration examples are provided in the supplementary material.


\noindent \textbf{Ablation Study}
To validate the effectiveness of the proposed hierarchical adapter and training strategy, we compare variants of \ourmethod.
Due to limited space, we use $\bullet$ to denote removing the CLIP branch in the first-stage training, $\star$ to denote random feature dropping for CLIP in the second stage, SA to denote single adapter using the last block in MoGe, and HA to denote \textit{4-layer} hierarchical adapter. 
Table \ref{tab:Srgpt} shows that single adapter generates SOTA results which validates the effectiveness of geometry-aware features in enhancing spatial grounding, but does not surpass hierarchical adapter on average.
In addition, we experiment with different numbers of adapter layers, where the 4-layer configuration significantly outperforms 3-layer (HA-3) and 5-layer (HA-5) alternatives.
We attribute this result to the adopted MoGe \cite{moge} network which relies on the features from the last four layers in the encoder for reconstruction.
Moreover, random CLIP feature dropping shows better results due to the emphasis on the MoGe module during training, while total removal of CLIP is not a feasible approach.


\begin{table}[t]
\centering
\caption{Evaluation on SpatialRGPT-Bench where the best results are marked in bold, and the second best results are underlined.}
\vspace{-0.2cm}
\resizebox{\columnwidth}{!}{ 
\begin{tabular}{lcccccc}
\toprule
\multirow{2}{*}{Model} & \multirow{2}{*}{Height} & \multirow{2}{*}{Width} & Vertical & Horizontal & Direct & \multirow{2}{*}{Average} \\ 
& & & Distance & Distance & Distance & \\ \midrule
LLaVA-1.5-7B & 10.53 & 15.04 & 16.98 & 17.21 & 13.51 & 14.49 \\
LLaVA-OSD & 54.14 & 34.59 & \textbf{56.60} & 50.82 & 40.54 & 46.73 \\
GPT-4o & 18.80 & 10.53 & 4.72 & 5.74 & 2.03 & 8.41 \\
GPT-4V & 24.06 & 21.05 & 6.60 & 9.02 & 7.43 & 13.86 \\
GPT-4.1 & \underline{61.65} & 36.84 & 2.83 & 9.02 & 19.59 & 27.10 \\
SpatialRGPT & \textbf{63.61} & \textbf{48.12} & 50.94 & 49.18 & 33.78 & 48.60 \\ \midrule
\multicolumn{3}{l}{\textit{Variants of \ourmethod}}\\
SpatialGeo-SA & 54.14 & \underline{44.36} & 54.72 & 55.74 & 38.51 & 48.91 \\
SpatialGeo-SA ($\star$) & 48.12 & 37.59 & \textbf{56.60} & \textbf{63.93} & \textbf{48.65} & \underline{50.47} \\
SpatialGeo-HA-3 ($\star$) & 41.35 & 38.34 & 53.77 & 45.08 & 41.21 & 43.45 \\
SpatialGeo-HA-5 ($\star$) & 39.85 & 37.59 & 50.00 & 38.52 & 37.84 & 40.34 \\
SpatialGeo-HA ($\bullet$,$\star$) & 18.05 & 17.29 & 26.42 & 18.85 & 21.62 & 20.25 \\ \midrule
\multicolumn{3}{l}{\textit{Full model of \ourmethod}} \\
SpatialGeo-HA ($\star$) & 58.65 & 41.35 & \textbf{56.60} & \underline{59.02} & \textbf{48.65} & \textbf{52.49} \\
\bottomrule
\end{tabular}
\label{tab:Srgpt}
}   
\vspace{-0.3cm}
\end{table}

\noindent \textbf{SpatialScore}
To evaluate the generalization of \ourmethod on other spatial reasoning benchmarks, we test on SpatialScore \cite{spatialscore}, which is a comprehensive multimodal spatial understanding benchmark covering various existing datasets.
We select those related to spatial reasoning as listed in Table \ref{tab:spatialScore} and select test samples that only contain a single image input.
As shown in Table \ref{tab:spatialScore}, \ourmethod excels the SOTA LLaVA-1.5-7B, especially on QSpatial-PluS where we have achieved a \textbf{51\%} improvement over it.
In addition, by comparing with its single-adapter variant, 
we show that \ourmethod with hierarchical adapter is more effective in spatial visual embedding.


\begin{table}[t]
    \centering
    \caption{Generalization Evaluation on SpatialScore.}
    \label{tab:spatialScore}
    \vspace{-0.2cm}
    \resizebox{\columnwidth}{!}{ 
    \begin{tabular}{lccc}
    \toprule
    &LLaVA-1.5-7B & SpatialGeo-SA($\star$) & SpatialGeo-HA($\star$) \\ \midrule
    QSpatial-Plus & 38.61 & 44.55 & \textbf{58.42} \\
    QSpatial-ScanNet & 47.65 & 50.59 & \textbf{54.12} \\
    SpatialBench & \textbf{53.45} & 51.15 & \textbf{53.45} \\
    VSR-ZeroShot & \textbf{70.13} & 69.31 & 68.66 \\
    SpatialSense & 60.24 & \textbf{63.20} & 63.11 \\
    RealWorldQA & 54.38 & \textbf{57.25} & 55.69 \\
    VGBench & 31.79 & 36.50 & \textbf{37.54} \\
    \midrule
    Average & 50.89 & 53.22  & \textbf{55.86} \\
    \bottomrule
    \end{tabular}
    }
    \vspace{-0.4cm}
\end{table}

\begin{table}[t]
    \centering
    \caption{Evaluation on General VQA Benchmarks.}
    \label{tab:public benchmarks}
    \vspace{-0.2cm}
    \resizebox{\columnwidth}{!}{ 
    \begin{tabular}{lccc}
    \toprule
    &LLaVA-1.5-7B & SpatialGeo-SA($\star$) & \multirow{1}{*}SpatialGeo-HA($\star$) \\ \midrule
    POPE (random)& \textbf{87.3} & \textbf{87.3} & 86.7 \\
    POPE (popular)  & \textbf{86.1} & 86.0 & 85.1 \\
    POPE (adversarial)  & 84.2 & \textbf{84.5} & 83.6 \\
    MM-Vet & \textbf{31.1} & 30.9 & 30.9 \\
    MME & \textbf{1504} & 1464 & 1470 \\
    MMVP  & 41 & 31 & \textbf{42} \\
    BLINK (RelD)
     & 54.64 & 68.04 & \textbf{73.20} \\
    \bottomrule
    \end{tabular}
    }
    \vspace{-0.5cm}
\end{table} 

\subsection{Evaluation on General VQA Benchmarks} 

We further test on general VQA benchmarks to evaluate whether the models overfit to the spatial tasks.
In particular, we include multiple benchmarks, \ie, POPE \cite{pope}, MM-Vet \cite{mm-vet}, MME \cite{2023mme}, MMVP \cite{mmvp} and BLINK (Relative Depth) \cite{blink} as listed in Table \ref{tab:public benchmarks}, and employ their respective evaluation criteria.
As shown in Table \ref{tab:public benchmarks}, \ourmethod performs similarly to LLaVA-1.5-7B baseline model. 
This result indicates that \ourmethod significantly \textit{improves its spatial reasoning ability without sacrificing its performance in general VQA tasks}.

It is worth noting that the BLINK \cite{blink} test set for relative depth evaluation is challenging as it evaluates the depth at the point level, which is not included in our training.
Despite that, \ourmethod still surpasses competing schemes by a large margin, validating its improvement in spatial reasoning.

\section{Conclusion}

We propose \ourmethod that aims at enhancing the spatial reasoning ability of MLLMs based on geometry-semantics feature fusion in the vision encoder. 
Specifically, we first analyze the visual shortcomings of existing MLLMs and unveil the lossy spatial encoding in their vision encoder CLIP.
Based on this analysis, we propose to integrate the geometry-aware MoGe encoder with LLaVA-1.5-7B foundation model to enhance spatial grounding of the model.
Then we design the hierarchical adapter to effectively fuse the geometry and semantic features without introducing extra 3D feature extraction modules.
To efficiently train the network, we use the pretrained LLaVA model to reduce training time and dataset requirement, then optimize the MoGe module with the random CLIP feature dropping strategy to enforce the utilization of MoGe features.


Extensive experiments validate that \ourmethod achieves SOTA performance in multiple spatial reasoning tasks and demonstrates strong generalization ability to unseen benchmarks.
The performance in general VQA benchmarks further validates that the spatial reasoning improvement of \ourmethod does not sacrifice its performance on general-purpose visual language tasks.

\bibliographystyle{IEEEtran}
\bibliography{refs}

\end{document}